\documentclass{article}

\usepackage{microtype}
\usepackage{graphicx}
\usepackage{subfigure}
\usepackage{booktabs} % for professional tables

\usepackage{hyperref}

\usepackage[accepted]{style_files/icml2023_deployableGenAI}

\usepackage{amsmath}
\usepackage{amssymb}
\usepackage{mathtools}
\usepackage{amsthm}

\usepackage[capitalize,noabbrev]{cleveref}

\theoremstyle{plain}

\theoremstyle{definition}

\theoremstyle{remark}

\usepackage{colortbl}
\usepackage{multirow}
\usepackage{makecell}
\usepackage{array}
\usepackage{hhline}
\newcommand{\internship}{$\text{}^{\ddagger}$Work conducted while interning at Recursion Pharmaceuticals.}
\usepackage[textsize=tiny]{todonotes}

\icmltitlerunning{Goal-conditioned GFNs for Controllable Molecular Design}

\begin{document}

\twocolumn[
\icmltitle{Goal-conditioned GFlowNets for Controllable Multi-Objective Molecular Design}



\icmlsetsymbol{intern}{$\ddagger$}

\begin{icmlauthorlist}
\icmlauthor{Julien Roy}{intern,mila,poly}
\icmlauthor{Pierre-Luc Bacon}{mila,udem,face}
\icmlauthor{Christopher Pal}{mila,poly,udem,sn,cifar}
\icmlauthor{Emmanuel Bengio}{rec}
\end{icmlauthorlist}

\icmlaffiliation{mila}{Mila.}
\icmlaffiliation{poly}{École Polytechnique de Montréal.}
\icmlaffiliation{udem}{Université de Montréal.}
\icmlaffiliation{sn}{ServiceNow.}
\icmlaffiliation{face}{Facebook CIFAR AI Chair.}
\icmlaffiliation{cifar}{Canada CIFAR AI Chair.}
\icmlaffiliation{rec}{Recursion Pharmaceuticals}

\icmlcorrespondingauthor{Julien Roy}{julien.roy@mila.quebec}

\icmlkeywords{Machine Learning, ICML}

\vskip 0.3in
]



\printAffiliationsAndNotice{\internship}

\begin{abstract}
In recent years, \textit{in-silico} molecular design has received much attention from the machine learning community. When designing a new compound for pharmaceutical applications, there are usually multiple properties of such molecules that need to be optimised: binding energy to the target, synthesizability, toxicity, EC50, and so on. While previous approaches have employed a scalarization scheme to turn the multi-objective problem into a \textit{preference-conditioned} single objective, it has been established that this kind of reduction may produce solutions that tend to slide towards the extreme points of the objective space when presented with a problem that exhibits a concave Pareto front. In this work we experiment with an alternative formulation of \textit{goal-conditioned} molecular generation to obtain a more controllable conditional model that can uniformly explore solutions along the entire Pareto front.
\end{abstract}

\section{Introduction}
\label{sec:introduction}

Modern Multi-Objective optimisation (MOO) is comprised of a large number of paradigms~\citep{keeney1993decisions,miettinen2012nonlinear} intended to solve the problem of trading off between different objectives; a setting particularly relevant to molecular design~\citep{jin2020multi,jain2022multi}. One particular paradigm that integrates well with recent discrete deep-learning based MOO is \emph{scalarization}~\citep{ehrgott2005multicriteria,pardalos2017non}, which transforms the problem of discovering the Pareto front of a problem into a \emph{family} of problems, each defined by a set of coefficients over the objectives. One notable issue with such approaches is that the solution they give tends to depend on the \emph{shape} of the Pareto front in objective space~\citep{emmerich2018tutorial}.

To tackle this problem, we propose to train models which explicitly target specific regions in \emph{objective space}. Taking inspiration from goal-conditional reinforcement learning~\citep{schaul2015universal}, we condition GFlowNet~\citep{bengio2021flow,bengio2021gflownet} models on a description of such \textit{goal regions}. Through the choice of distribution over these goals, we enable users of these models to have more fine-grained control over trade-offs. We also find that assuming proper coverage of the goal distribution, goal-conditioned models discover a more complete and higher entropy approximation of the Pareto front for various shapes.

\section{Background \& Related Work}
\label{sec:related_work}

The \textbf{Multi-Objective optimisation} problem can be broadly described as the desire to maximize a set of $K$ objectives over $\mathcal{X}$, $\mathbf{R}(x)\in\mathbb{R}^K$. In typical MOO problems, there is no single optimal solution $x$ such that $R_k(x)>R_k(x') \, \forall \, k, x'$. Instead, the solution set is generally composed of \emph{Pareto optimal} points, which are points $x$ that are not \emph{dominated} by any other point, i.e. $\nexists \, x'\mbox{ s.t. }R_k(x)\geq R_k(x') \, \forall \, k$. In other words, a point is Pareto optimal if it cannot be locally improved. The projection in objective space of the set of Pareto optimal points forms the so-called \emph{Pareto front}. 

\begin{figure*}[h!]
    \centering
    \centering
    \begin{minipage}{.33\textwidth}
        \centering
        \includegraphics[height=4cm]{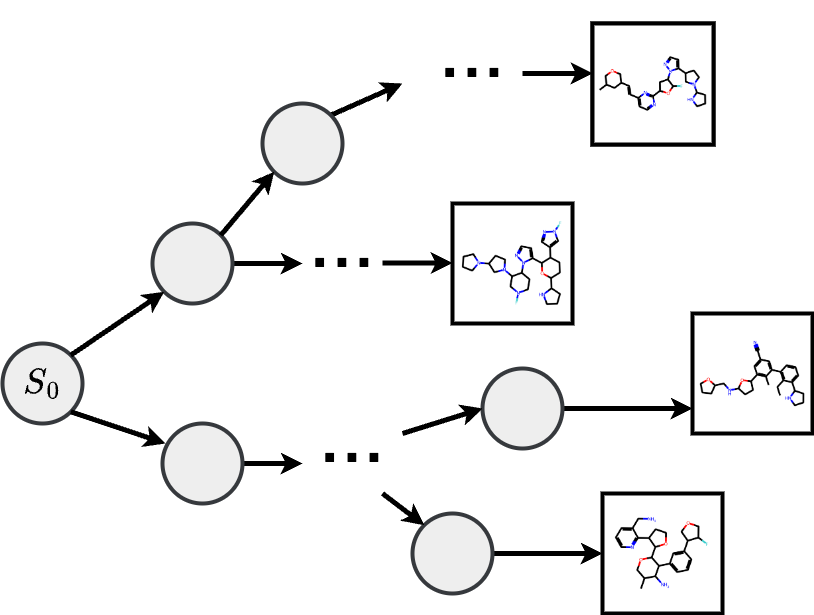}
    \end{minipage}%
    \begin{minipage}{0.66\textwidth}
        \centering
        \includegraphics[height=4cm]{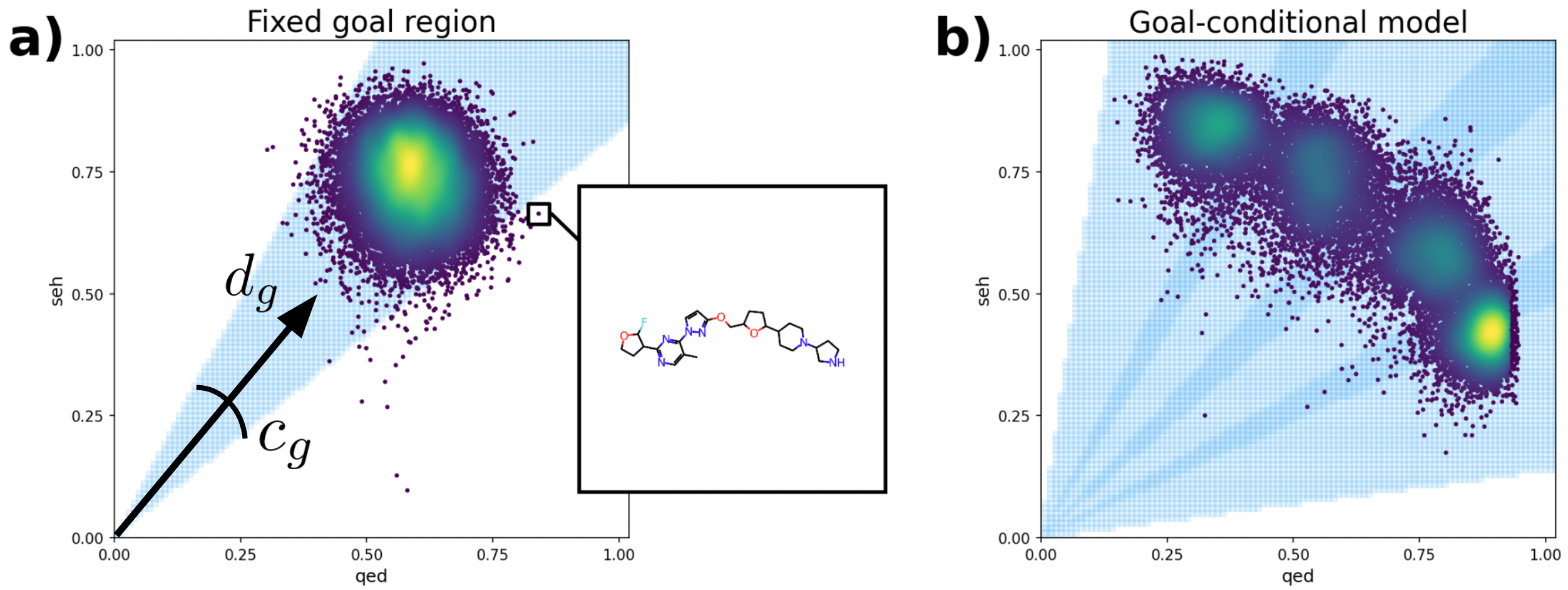}
    \end{minipage}
    \caption{The diagram on the left depicts the state space of a GFlowNet molecule generator which learns a forward policy that sequentially builds diverse molecules. \textbf{a)} The sampling distribution learned by such a model on a two-objective problem (seh, qed). Each dot represents a molecule's image in the objective space. The focus region (see Section~\ref{sec:goal-conditioned-gfn}) is depicted as a light blue cone, and the colors represent the density of the distribution. The model learns to produce molecules that mostly belong within the focus region. \textbf{b)} By training a goal-conditioned GFlowNet and sampling from several focus regions (here showing 4 distinct regions), we can cover a wider section of the objective space and increase the diversity of proposed candidates.}
    \label{fig:method_overview}
\end{figure*}

As graph-based models improve~\citep{rampavsek2022recipe} and more molecular data become available~\citep{wu2018moleculenet}, molecular design has become an active field of research within the deep learning community~\citep{brown2019guacamol,huang2021therapeutics}, and core to this research is the fact that molecular design is a fundamentally multi-objective search problem~\citep{papadopoulos2006multiobjective,brown2006novel}. The advent of such tools has led to various important work at the intersection of these two fields~\citep{zhou2019optimization,staahl2019deep,jin2020multi,jain2022multi}.

The \textbf{Generative Flow Network} (GFlowNet, GFN) framework is a recently introduced method to train energy-based generative models ~\citep[i.e. models that learn $p_\theta(x) \propto R(x)$; ][]{bengio2021flow}. They have now been successfully applied to a variety of settings such as biological sequences~\citep{jain2022biological}, causal discovery~\citep{deleu2022bayesian,atanackovic2023dyngfn}, discrete latent variable modeling~\citep{hu2023gflownet}, and computational graph scheduling~\citep{zhang2023robust}. The framework itself has also received theoretical attention~\citep{bengio2021gflownet}, for example, highlighting its connections to variational methods~\citep{zhang2022unifying,malkin2022gflownets}, and several objectives to train GFNs have been proposed~\citep{malkin2022trajectory,madan2022learning,pan2023better} including extensions to continuous domains~\citep{lahlou2023theory}. 

In the context of molecular design, GFlowNets have several important properties that make them an interesting method for this task. Notably, they are naturally well-suited for discrete compositional object generation, and their multi-modal modeling capabilities allow them to induce greater state space diversity in the solutions they find than previous methods. A recent GFN-based approach to multi-objective molecular design, which we call \textit{preference-conditioning}~\cite{jain2022multi}, amounts to scalarizing the objective function by using a set of weights (or preferences) $w$:
\begin{equation}
R_w(x) = \sum_k w_k r_k \quad , \quad \sum_k w_k = 1 \quad , \quad w_k \geq 0
\end{equation}
and then passing this preference vector $w$ as input to the model. By sampling various $w$'s from a distribution such as Dirichlet's during training, one can obtain a model that can be conditioned to emphasize some preferred dimensions of the reward function. \citet{jain2022multi} also find that such a method finds diverse candidates in both state and objective spaces.

\section{Methods}
\label{sec:methods}

\subsection{Goal-conditioned GFlowNets}
\label{sec:goal-conditioned-gfn}

Building on the method of \citet{jain2022multi}, our approach also formulates the problem as a conditional generative task but now imposes a hard constraint on the model: the goal is to generate samples for which the image in objective space falls into the specified goal region. While many different goal-design strategies could be employed, we take inspiration from \citet{lin2019pareto} and state that a sample $x$ meets the specified goal $g$ if the cosine similarity between its reward vector $r$ and the goal direction $d_g$ is above the threshold $c_g$: $g := \{r \in \mathbb{R}^K: \frac{r \cdot d_g}{||r||\cdot||d_g||} \geq c_g \}$. We call such a goal a \textit{focus region}, which represents a particular choice of trade-off in the objective space (see Figure~\ref{fig:method_overview}). The method can be considered a form of goal-conditional reinforcement learning~\cite{schaul2015universal}, where the reward function $R_g$ depends on the current goal $g$. In our case we have:
\begin{equation}
\label{eq:goal-reward}
    R_g(x) = 
    \begin{cases}
    \sum_k r_k ,& \text{if } r \in g\\
    0,              & \text{otherwise}
    \end{cases}
\end{equation}
To alleviate the effects of the now increased sparsity of the reward function $R_g$, we use a replay buffer which proved to stabilise the learning dynamics of our models (see Appendix~\ref{app:replay_buffer}). Notably, by explicitly formulating a goal, we can measure the \textit{goal-reaching accuracy} of our model, which refers to the proportion of samples that successfully landed in their prescribed region. This measurement enables us to employ hindsight experience replay~\citep{andrychowicz2017hindsight}, which lets the model learn from the sampled trajectories that didn't meet their goal. Finally, to further increase the goal-reaching accuracy we sharpen the reward function's profile to help the model generate samples closer to the center of the focus region (see Appendix~\ref{app:focus_limit_coef}). 

\subsection{Learned Goal Distribution}
\label{sec:learned_goal_distribution}

Preference conditioning uses soft constraints to steer the model in some regions of the objective space. While hard constraints provide a more explicit way of incorporating the user's intentions in the model~\citep{amodei2016concrete, roy2021direct}, they come with the unique challenge that not every goal may be feasible. In such cases, the model will only observe samples with a reward of 0 and thus return molecules of little interest drawn uniformly across the state space. These ``bad samples" are not harmful in themselves and can easily be filtered out. Still, their prominence will affect the sampling efficiency of goal-conditioned approaches compared to their soft-constrained counterpart. Moreover, the number of infeasible regions will likely multiply as the number of objectives grows, further aggravating this disparity. To cope with this challenge, we propose to use a simple \textit{tabular goal-sampler} (Tab-GS) which maintains a belief about whether any particular goal direction $d_g$ is feasible. Once learned, we can start drawing new goals from it with a much lower likelihood on the goals that are believed to be infeasible, thus restoring most of the lost sample efficiency. We give more details on this approach in Appendix~\ref{app:learned_goal_model} and use it in our experiments in Section~\ref{sec:increasing_number_of_objectives}.

% First result package
\begin{figure*}[h!]
    \centering
    \includegraphics[width=1\linewidth, height=2.75cm]{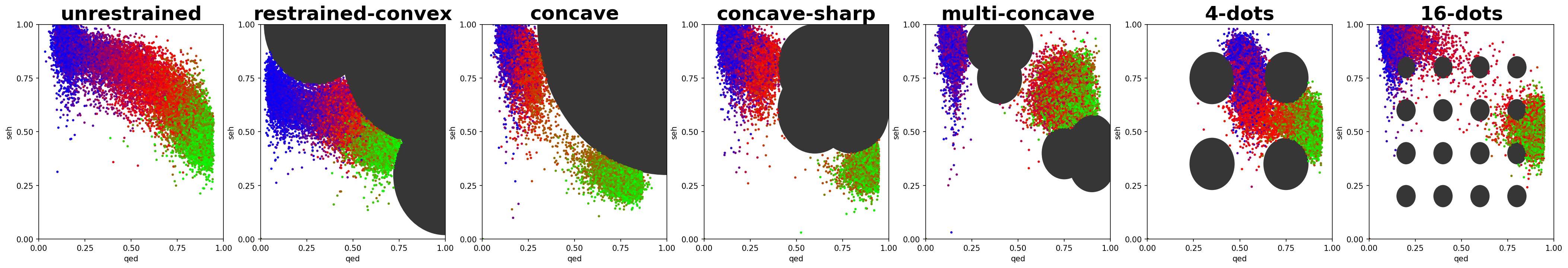}
    \includegraphics[width=1\linewidth, height=2.75cm]{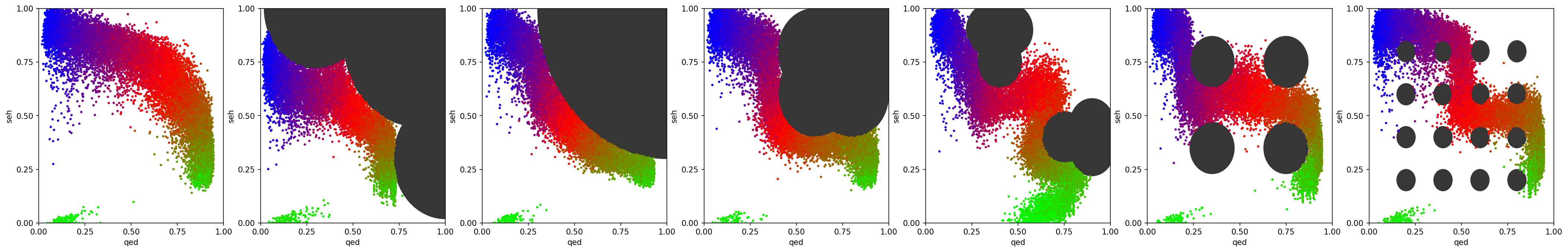}
    \caption{Comparisons between a preference-conditioned GFN (top row) and a goal-conditioned GFN (bottom row) on a set of increasingly complex modifications of a two-objective (seh, qed) fragment-based molecule generation task~\cite{jain2022multi}. The BRG colors represent the angle between the vector $[1, 0]$ and either the preference-vector $w$ (top) or the goal direction $d_g$ (bottom), respectively. For example, in the case of preference-conditioning, a green dot means that such samples were produced with a strong preference for the qed-objective, while in the goal-conditioning case, a green dot means that  the model \textit{intended} to produce a sample alongside the qed-axis. We see that goal-conditioning allows to span the entire objective space even in very challenging landscapes (columns 3-7) and in a more controllable way.}
    \label{fig:difficult_pareto_fronts_alignment}
\end{figure*}
\begin{figure*}[h!]
    \centering
    \includegraphics[width=1\linewidth, height=2.75cm]{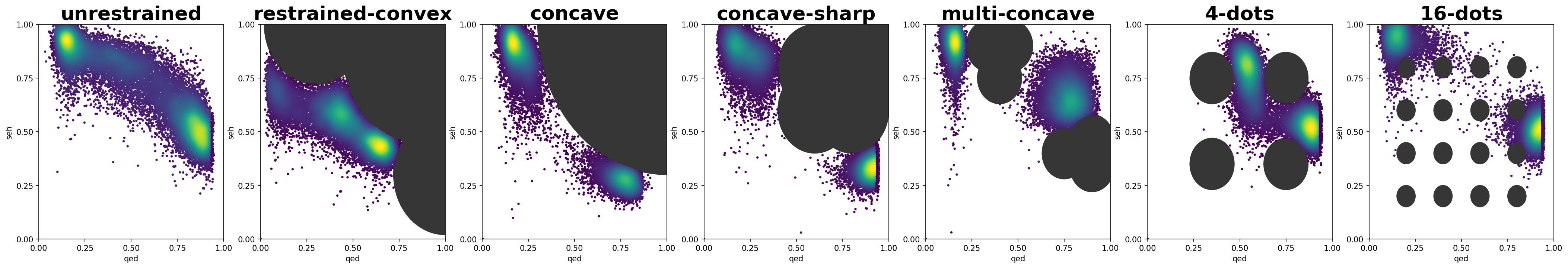}
    \includegraphics[width=1\linewidth, height=2.75cm]{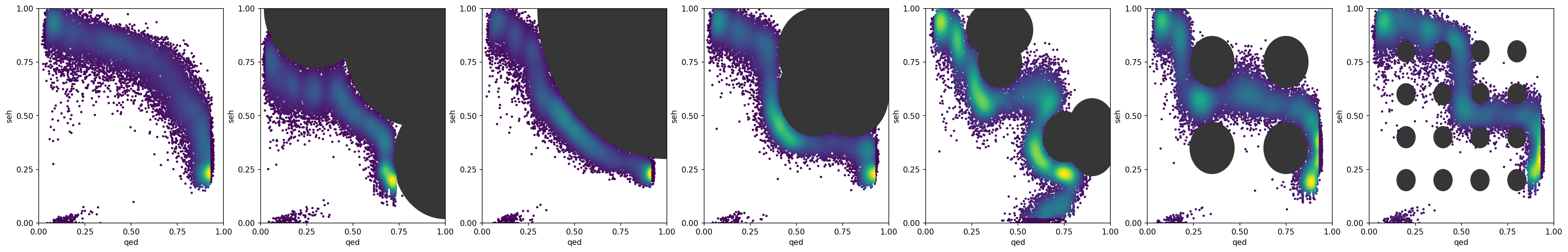}
    \caption{Comparisons of the same sampling distributions depicted in Figure~\ref{fig:difficult_pareto_fronts_alignment}. Now the colors indicate how densely populated a particular area of the objective space is (brighter is more populated). We can see that by explicitly targeting different trade-off regions in objective space, our goal-conditioning approach (bottom row) produces far more evenly distributed samples along the Pareto front than with preference-conditioning (top row).}
    \label{fig:difficult_pareto_fronts_density}
\end{figure*}

\begin{table*}[h!]
    \centering
    \caption{Comparisons according to IGD, Avg-PCC and PC-ent between preference-conditioned and goal-conditioned GFNs on a set of increasingly difficult objective landscapes, metrics reported on 3 seeds (mean $\pm$ sem).}
    \vspace{2mm}
    \renewcommand{\arraystretch}{1.3}
    \resizebox{\textwidth}{!}{ 
    \begin{tabular}{|r|c|c c c c c c c|}
        \cline{2-9}
         \multicolumn{1}{c|}{} & \textbf{algorithm} & \textbf{unrestrained} & \textbf{restrained-convex} & \textbf{concave} & \textbf{concave-sharp} & \textbf{multi-concave} & \textbf{4-dots} & \textbf{16-dots}\\
        \hline
        \multirow{2}{*}{IGD $(\downarrow)$} & pref-cond & \cellcolor{lightgray}$\mathbf{0.087 \pm 0.001}$ & $0.316 \pm 0.002$ & $0.272 \pm 0.001$ & \cellcolor{lightgray}$\mathbf{0.180 \pm 0.002}$ & \cellcolor{lightgray}$\mathbf{0.152 \pm 0.006}$ & $0.130 \pm 0.011$ & $0.109 \pm 0.009$ \\
        \cline{2-9}
         & goal-cond & $0.095 \pm 0.002$ & \cellcolor{lightgray}$\mathbf{0.310 \pm 0.001}$ & \cellcolor{lightgray}$\mathbf{0.266 \pm 0.001}$ & $0.197 \pm 0.002$ & $0.173 \pm 0.004$ & $0.134 \pm 0.002$ & $0.115 \pm 0.004$ \\
        \hline
        \hline
        \multirow{2}{*}{Avg-PCC $(\uparrow)$} & pref-cond & $0.905 \pm 0.001$ & $0.673 \pm 0.009$ & $0.830 \pm 0.002$ & $0.855 \pm 0.004$ & $0.700 \pm 0.009$ & $0.768 \pm 0.038$ & $0.770 \pm 0.011$ \\
        \cline{2-9}
         & goal-cond & \cellcolor{lightgray}$\mathbf{0.967 \pm 0.002}$ & \cellcolor{lightgray}$\mathbf{0.953 \pm 0.001}$ & \cellcolor{lightgray}$\mathbf{0.926 \pm 0.002}$ & \cellcolor{lightgray}$\mathbf{0.915 \pm 0.001}$ & \cellcolor{lightgray}$\mathbf{0.946 \pm 0.004}$ & \cellcolor{lightgray}$\mathbf{0.928 \pm 0.002}$ & \cellcolor{lightgray}$\mathbf{0.948 \pm 0.001}$ \\
        \hline
        \hline
        \multirow{2}{*}{PC-ent $(\uparrow)$} & pref-cond & $2.170 \pm 0.004$ & $1.913 \pm 0.019$ & $1.563 \pm 0.009$ & $1.629 \pm 0.002$ & $1.867 \pm 0.015$ & $1.521 \pm 0.022$ & $1.610 \pm 0.019$ \\
        \cline{2-9}
         & goal-cond & \cellcolor{lightgray}$\mathbf{2.472 \pm 0.006}$ & \cellcolor{lightgray}$\mathbf{2.242 \pm 0.013}$ & \cellcolor{lightgray}$\mathbf{1.997 \pm 0.002}$ & \cellcolor{lightgray}$\mathbf{1.918 \pm 0.001}$ & \cellcolor{lightgray}$\mathbf{2.380 \pm 0.020}$ & \cellcolor{lightgray}$\mathbf{2.270 \pm 0.025}$ & \cellcolor{lightgray}$\mathbf{2.262 \pm 0.014}$ \\
        \hline
    \end{tabular}
    }
\label{tab:difficult_pareto_fronts}
\end{table*}

\subsection{Evaluation Metrics}
\label{sec:evaluation_metrics}

While there exists many multi-objective scoring functions to choose from, any single metric only partially captures the desirable properties of the learned generative distribution~\citep{audet2021performance}. In this work, we focus on sampling high-performing molecules across the entire Pareto front in a controllable manner at test time. With that in mind, we propose combining three metrics to evaluate our solution. The first one, the Inverted Generational Distance (IGD)~\citep{coello2005solving}, uses a set of reference points $P$ (the \textit{true} Pareto front) and takes the average of the distance to the closest generated sample for each of these points:
$\text{IGD}(S, P) := \frac{1}{|P|} \sum_{p \in P} \min_{s \in S} ||s - p||_2^2$ where $S=\{s_i\}_{i=1}^N$ is the image in objective space of a set of $N$ generated molecules $s_i$. When the true Pareto front is unknown, we use a discretization of the extreme faces of the objective space hypercube as reference points. IGD thus captures the width and depth at which our Pareto front approximation reaches out in the objective space. The second metric, which we call the Pareto-Clusters Entropy (PC-ent), measures how uniformly distributed the samples are along the true Pareto front. To accomplish this, we use the same reference points $P$ as for IGD, and cluster together in the subset $S_j$ all of the samples $s_i$ located closer to the reference point $p_j$ than any other reference point. PC-ent computes the entropy of the histogram of each counts $|S_j|$, reaching its maximum value of $-\log \frac{1}{|P|}$ when all the samples are uniformly distributed relative to the true Pareto front:
$\text{PC-ent}(S, P) := - \sum_{j} \frac{|S_j|}{|P|} \log \frac{|S_j|}{|P|}$. Finally, to report on the \textit{controllability} of the compared methods, we measure the Pearson correlation coefficient (PCC) between the conditional vector $c$ (goal or preference) and the resulting reward vector $s$, averaged across objectives $k$:
$\text{Avg-PCC}(S, C) := \frac{1}{K}\sum_{k=1}^K\text{PCC}(s_{\cdot,k}, c_{\cdot,k})$.

\section{Results}
\label{sec:results}

\subsection{Evaluation Tasks}
\label{sec:task}

We primarily experiment on a two-objective task, the well-known drug-likeness heuristic QED~\citep{bickerton2012quantifying}, which is already between 0 and 1, and the sEH binding energy prediction of a pre-trained publicly available model~\citep{bengio2021flow}; we divide the output of this model by 8 to ensure it will likely fall between 0 and 1 (some training data goes past values of 8). For 3 and 4 objective tasks, we use a standard heuristic of synthetic accessibility~\citep{ertl2009estimation} and a penalty for compounds exceeding a molecular weight of 300. See Appendix~\ref{app:training_details} for all task and training details.

\subsection{Comparisons in Difficult Objective Landscapes}
\label{sec:difficult_landscapes}

To simulate the effect of complexifying the objective landscape while keeping every other parameter of the evaluation fixed, we incorporate \textit{unreachable regions}, depicted in dark in Figures~\ref{fig:difficult_pareto_fronts_alignment}~\&~\ref{fig:difficult_pareto_fronts_density}, by simply setting to \textit{null} the reward function of any molecule whose image in the objective space would fall into these dark regions. We can see that the preference-conditioned approach can effectively solve problems exhibiting a convex pareto-front (Figure~\ref{fig:difficult_pareto_fronts_alignment}~\&~\ref{fig:difficult_pareto_fronts_density}, columns 1-2). However, it is far less effective on problems exhibiting more complex objective landscapes. When faced with a concave Pareto front, the algorithm favours solutions towards the extreme ends (Figure~\ref{fig:difficult_pareto_fronts_alignment}~\&~\ref{fig:difficult_pareto_fronts_density}, columns 3-7). In contrast, by explicitly forcing the algorithm to sample from each trade-off direction in the objective space, our goal-conditioned method learns a sampling distribution that spans the entire space diagonally, no matter how complex we make the objective landscape. Table~\ref{tab:difficult_pareto_fronts} reports the performance of both methods on these objective landscapes in terms of IGD, Avg-PCC and PC-ent (mean $\pm$ sem, over 3 seeds). We see in Table~\ref{tab:difficult_pareto_fronts} that according to IGD, preference-conditioning and goal-conditioning perform similarly in terms of pushing the empirical Pareto front forward. While the two learned distributions are in many cases very different (Figure~\ref{fig:difficult_pareto_fronts_alignment}, columns 3-7), the preference-conditioning method still manages to produce a few samples in the middle areas of the Pareto front, which satisfies IGD as it only looks for the single closest sample to each reference point. However, the two algorithms differ drastically in terms of controllability of the distribution (color-coded in  Figure~\ref{fig:difficult_pareto_fronts_alignment}) and uniformity of the distribution along the Pareto front (color-coded in  Figure~\ref{fig:difficult_pareto_fronts_density}), which are highlighted by the Avg-PCC and PC-ent criteria in Table~\ref{tab:difficult_pareto_fronts}.

\subsection{Comparisons for Increasing Number of Objectives}
\label{sec:increasing_number_of_objectives}

Using the same metrics, we also evaluate the performance of both methods when the number of objectives increases. As described in Section~\ref{sec:learned_goal_distribution}, to maintain the sample efficiency of our goal-conditioned approach we sample the goal directions $d_g$ from a learned tabular goal-sampler (Tab-GS) rather than uniformly across the objective space (Uniform-GS). We can see in Table~\ref{tab:growing_number_of_objectives} (and Appendix~\ref{app:learned_goal_model}) that with this adaptation, our goal-conditioned approach maintains its advantages in terms of controllability and uniformity of the learned distribution as the number of objectives increases, proving to be an effective method for probing large, high-dimensional objective spaces for diverse solutions.

\begin{table}[h!]
    \centering
    \caption{Comparisons according to IGD, Avg-PCC and PC-ent between preference- and goal-conditioned GFNs faced with increasing objectives (3 seeds, mean $\pm$ sem).}
    \vspace{1mm}
    \renewcommand{\arraystretch}{1.3}
    \resizebox{\linewidth}{!}{ 
    \begin{tabular}{|r|c|c c c|}
        \cline{2-5}
         \multicolumn{1}{c|}{} & \textbf{algorithm} & \textbf{2 objectives} & \textbf{3 objectives} & \textbf{4 objectives}\\
        \hline
        \multirow{2}{*}{IGD $(\downarrow)$} & pref-cond & \cellcolor{lightgray}$\mathbf{0.088 \pm 0.001}$ & $0.218 \pm 0.003$ & $0.370 \pm 0.000$ \\
        \cline{2-5}
         & goal-cond & $0.094 \pm 0.004$ & \cellcolor{lightgray}$\mathbf{0.199 \pm 0.002}$ & \cellcolor{lightgray}$\mathbf{0.303 \pm 0.001}$ \\
        \hline
        \hline
        \multirow{2}{*}{Avg-PCC $(\uparrow)$} & pref-cond & $0.904 \pm 0.002$ & $0.775 \pm 0.004$ & $0.612 \pm 0.002$ \\
        \cline{2-5}
         & goal-cond & \cellcolor{lightgray}$\mathbf{0.961 \pm 0.001}$ & \cellcolor{lightgray}$\mathbf{0.909 \pm 0.001}$ & \cellcolor{lightgray}$\mathbf{0.893 \pm 0.002}$ \\
        \hline
        \hline
        \multirow{2}{*}{PC-ent $(\uparrow)$} & pref-cond & $2.166 \pm 0.007$ & $3.775 \pm 0.016$ & $4.734 \pm 0.004$ \\
        \cline{2-5}
         & goal-cond & \cellcolor{lightgray}$\mathbf{2.471 \pm 0.001}$ & \cellcolor{lightgray}$\mathbf{4.571 \pm 0.008}$ & \cellcolor{lightgray}$\mathbf{6.320 \pm 0.009}$ \\
        \hline
    \end{tabular}
    }
\label{tab:growing_number_of_objectives}
\end{table}

\section{Future Work}
\label{sec:discussion}
\begin{figure}[b]
    \centering
    \vspace{-4mm}
    \includegraphics[width=\linewidth, height=3cm]{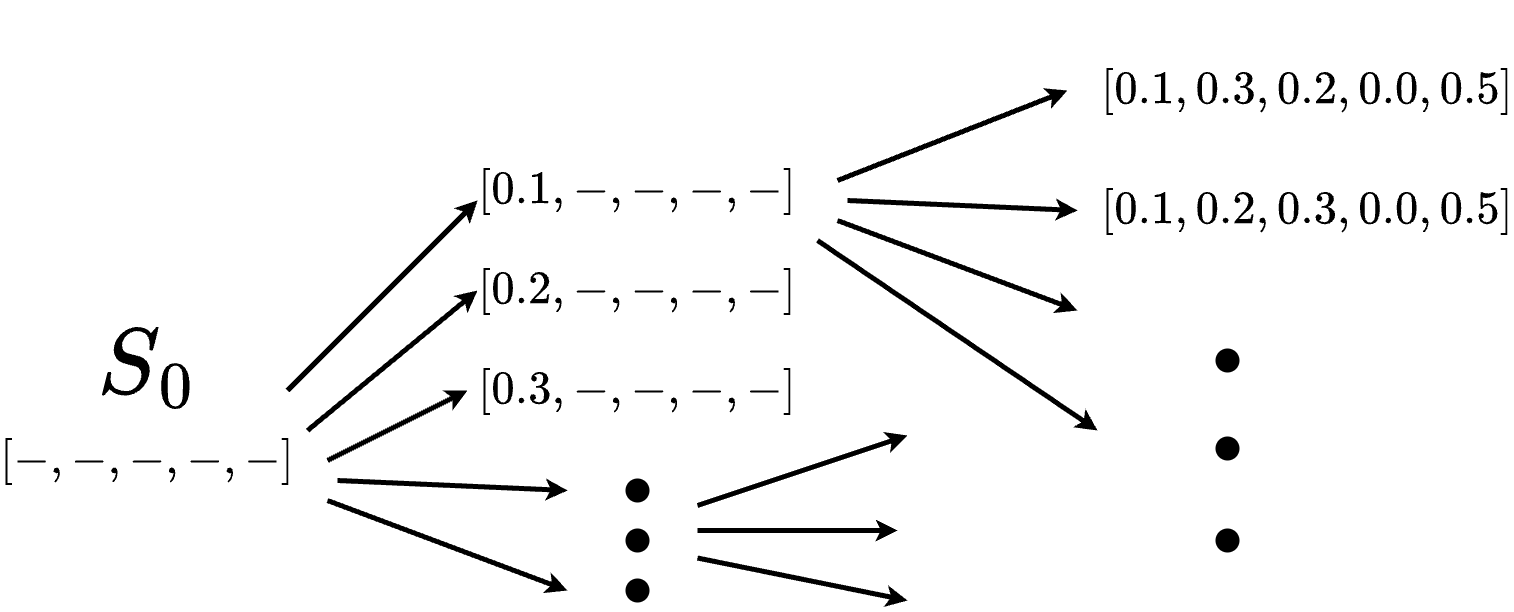}
    \caption{Depiction of a GFlowNet Goal Sampler (GFN-GS) gradually building goal directions $d_g$ as a sequence of $K$ steps.}
    \label{fig:hirarchical_goal_sampler}
\end{figure}
In this work, we proposed goal-conditioned GFlowNets for multi-objective molecular design. We showed that they are an effective solution to give practitioners more control over their generative models, allowing them to obtain a large set of more widely and more uniformly distributed molecules across the objective space. An important limitation of the proposed approach was the reduced sample efficiency of the method due to the existence of \textit{a priori unknown} infeasible goals. We proposed a tabular approach to gradually discredit these fruitless goal regions as we explore the objective space. However, this set of parameters, one for every goal direction, grows exponentially with the number of objectives $K$, eventually leading to statistical and memory limitations. As future steps, we plan to experiment with a GFlowNet-based Goal Sampler (GFN-GS) which would learn to sample feasible goal directions dimension by dimension, thus benefiting from parameter sharing and the improved statistical efficiency of its hierarchical structure.

\section*{Acknowledgements}

We wish to thank Berton Earnshaw, Paul Barde and Tristan Deleu as well as the entire research team at Recursion's Emerging ML Lab for providing insightful comments on this work. We also acknowledge funding in support of this work from Fonds de Recherche Nature et Technologies (FRQNT), Institut de valorisation des données (IVADO) and Recursion Pharmaceuticals.


% \clearpage
\bibliography{sources}
\bibliographystyle{style_files/icml2023}

%%%%%%%%%%%%%%%%%%%%%%%%%%%%%%%%%%%%%%%%%%%%%%%%%%%%%%%%%%%%%%%%%%%%%%%%%%%%%%%
%%%%%%%%%%%%%%%%%%%%%%%%%%%%%%%%%%%%%%%%%%%%%%%%%%%%%%%%%%%%%%%%%%%%%%%%%%%%%%%
% APPENDIX
%%%%%%%%%%%%%%%%%%%%%%%%%%%%%%%%%%%%%%%%%%%%%%%%%%%%%%%%%%%%%%%%%%%%%%%%%%%%%%%
%%%%%%%%%%%%%%%%%%%%%%%%%%%%%%%%%%%%%%%%%%%%%%%%%%%%%%%%%%%%%%%%%%%%%%%%%%%%%%%
\clearpage
\appendix
\onecolumn

\clearpage
\section{Task and Training Details}
\label{app:training_details}

We use the GFlowNet framework~\cite{bengio2021flow, bengio2021gflownet} to train discrete distribution samplers over the space of molecules that can be assembled from a set of pre-defined molecular fragments~\cite{kumar2012fragment}. A state is represented as a graph in which each node represents a fragment from the fragment library and where each edge has two attributes representing the attachment point of each connected fragment to its neighbor. The state representation is augmented with a fully-connected virtual node, whose features are an embedding of the conditioning information computed from the conditioning vector that represents the preferences $w$ and/or the goal direction $d_g$. To produce the state-conditional distribution over actions, the model processes the state using a graph transformer architecture~\cite{yun2019graph} for a predefined number of message-passing steps (number of layers). Our GFlowNet sampler thus starts from the initial state $s_0$ representing an empty graph. It iteratively constructs a molecule by either adding a node or an edge to the current state $s_t$ until it eventually selects the `STOP' action. 

To maintain some amount of exploration throughout training, at each construction step $t$, the model samples a random action with probability $\epsilon$ and otherwise samples from its forward transition distribution. The model is trained using the trajectory balance criterion~\cite{malkin2022trajectory} and thus is parameterised by a forward action distribution $P_F$ and an estimation of the partition function $Z:= \sum_x R(x)$. Forbidden actions are masked out from the forward transition distribution (for example, the action of adding an edge to the empty state). We use a uniform distribution for the backward policy $P_B$. To prevent the sampling distribution from changing too abruptly, we collect new trajectories from a sampling model $P_F(\, \cdot \,|\theta_{\text{sampling}})$ which uses a soft update with hyperparameter $\tau$ to track the learned GFN at update $k$: $\theta_{\text{sampling}}^{(k)} \leftarrow \tau \cdot \theta_{\text{sampling}}^{(k-1)} + (1 - \tau) \cdot \theta^{(k)}$. This is akin to the target Q-functions and target policies used in actor-critic frameworks~\cite{mnih2015human, fujimoto2018addressing}. 

The hyperparameters used for training both methods are listed in Table~\ref{tab:hyperparameters}.

\begin{table}[h!]
    \centering
    \caption{Hyperparameters used in our conditional-GFN training pipeline}
    \scriptsize
    \renewcommand{\arraystretch}{1.3}
    \begin{tabular}{|l|c|c|}
        \hline
        \multirow{2}{*}{\textbf{Hyperparameters}} & \multicolumn{2}{c|}{\textbf{Values}} \\ 
        \cline{2-3} & \textbf{Goal-conditioned GFN} & \textbf{Preference-conditioned GFN} \\ 
        \hline
        Batch size & 64 & 64 \\
        GFN temperature parameter $\beta$ & 60 & 60 \\
        Number of training steps & 40,000 & 40,000 \\
        Number of GNN layers & 2 & 2 \\
        GNN node embedding size & 256 & 256 \\
        Learning rate for GFN's $P_F$ & $10^{-4}$ & $10^{-4}$ \\
        Learning rate for GFN's $Z$-estimator & $10^{-3}$ & $10^{-3}$ \\
        Sampling moving average $\tau$ & 0.95 & 0.95 \\
        Random action probability $\epsilon$ & 0.01 & 0.01 \\
        Focus region cosine similarity threshold $c_g$ & 0.98 & - \\
        Limit reward coefficient $m_g$ & 0.20 & - \\
        Replay buffer length & 100,000 & - \\
        Number of replay buffer trajectory warmups & 1,000 & - \\
        Hindsight ratio & 0.30 & - \\
        Conditioning-vector sampling distribution &
        $d_g \sim 
        \begin{cases}
        \text{Uniform-GS} &\text{ (Sec~\ref{sec:difficult_landscapes})} \\
        \text{Tab-GS} &\text{ (Sec~\ref{sec:increasing_number_of_objectives})}
        \end{cases}$ & $w \sim Dirichlet(1)$ \\
        \rule{0pt}{1pt} & & \\
        \hline
        \end{tabular}
        \label{tab:hyperparameters}
\end{table}

\clearpage
\section{Failure Modes and Filtering}
\begin{figure}[ht]
    \centering
    \includegraphics[width=0.9\textwidth]{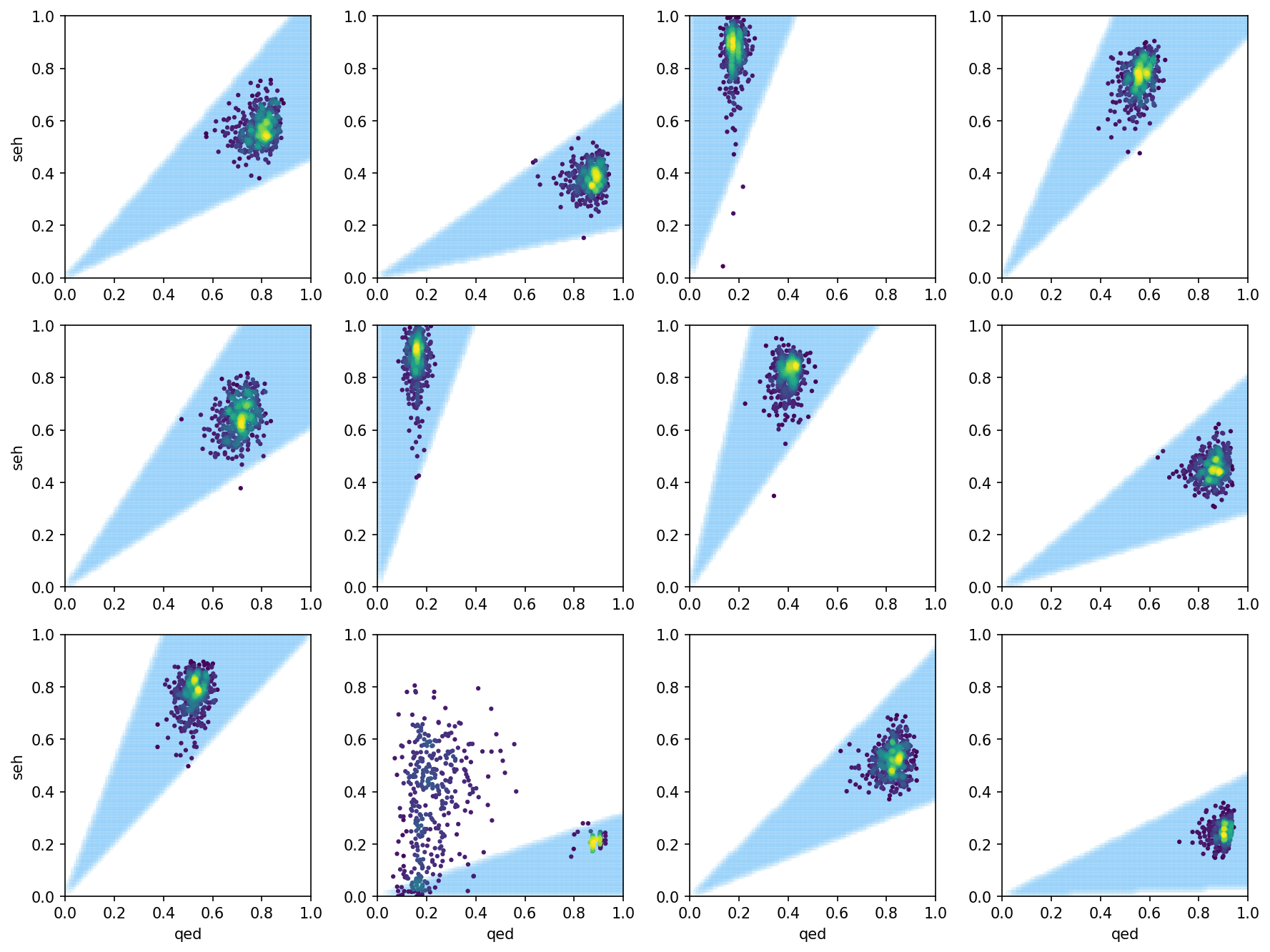}
    \caption{Learned conditional-distributions for different focus regions passed as input to the same model. Each dot marks the image of a generated molecule in the objective space. The colors indicate how densely populated a particular area of the objective space is (brighter is denser). The focus regions (goal regions) are depicted in light blue. The distribution on the last row, second column, showcases a focus region which seems difficult to reach and may not contain as large a population of molecules in the state space. In such cases, the model cannot learn to consistently produce samples from that goal region when conditioned on this goal direction $d_g$ and will instead produce several samples very similar to the sampling distribution of an untrained model (uniform across the state space).}
    \label{fig:goal_conditioning}
\end{figure}

While using goal regions as hard constraints offers a more precise tool for controllable generation, it faces the additional challenge that not all goals may be feasible (or that reaching some goals may be much easier to learn than others). When a model is conditioned with an infeasible goal, all the samples that it will observe will have a reward $R(x)=0$. The proper behavior, in that case, is to sample any possible molecule with equal weight, thus sampling uniformly across the entire molecular state space. Such molecules generally won't be of any interest and can be discarded. Thus, in our experiments, we filter out such \textit{out-of-focus} samples (molecules falling outside the focus region) and evaluate the candidates that were inside their prescribed focus region. Figure~\ref{fig:goal_conditioning} shows the conditional distributions learned by a single model trained on the 2-objective task. The picture on the last row, second column showcases such an occurrence of difficult focus region which results in many samples simply belonging to the uniform distribution over the state space.

\clearpage
\section{Ablations}
\label{app:ablations}
\subsection{Replay Buffer}
\label{app:replay_buffer}

\begin{figure}[h!]
    \centering
    \includegraphics[width=1.\textwidth]{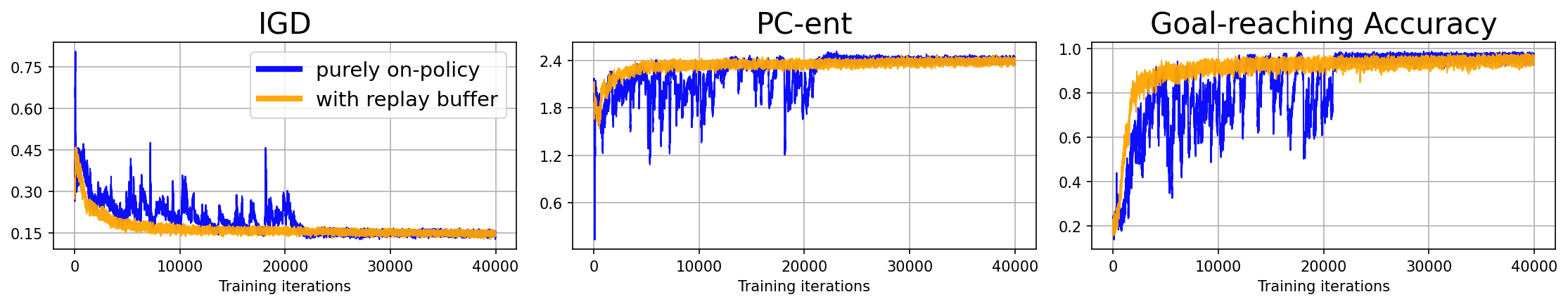}
    \caption{Learning curves for goal-conditioned models either trained purely on-policy (in blue) or using a replay buffer of past trajectories (in orange) on the 2-objective (seh, qed) task.}
    \label{fig:replay_buffer_effect}
\end{figure}

While both the un-conditional and the preference-conditioned GFN models are learning stably even in a purely on-policy setting, we found that the goal-conditioned models were more prone to instabilities and mode-collapse when employed purely on-policy (see Figure~\ref{fig:replay_buffer_effect}). This could be because imposing these hard constraints on the generative behavior of the model drastically changes the reward landscape from one set of goals to another. While larger batches could potentially alleviate this problem, sampling uniformly from a replay buffer of the last trajectories proved effective, as observed in many works stemming from \citet{mnih2015human}. 
As described in Section~\ref{sec:goal-conditioned-gfn}, we also use hindsight experience replay~\citep{andrychowicz2017hindsight}. Specifically, for every batch of data, we randomly select a subset of trajectories (hindsight-ratio * batch-size), among which we re-label both the goal direction $d_g$ and the corresponding reward for the examples that didn't reach their goal. 

\clearpage
\subsection{Limit Reward Coefficient}
\label{app:focus_limit_coef}

\begin{figure}[h!]
    \centering
    \includegraphics[width=1.\textwidth]{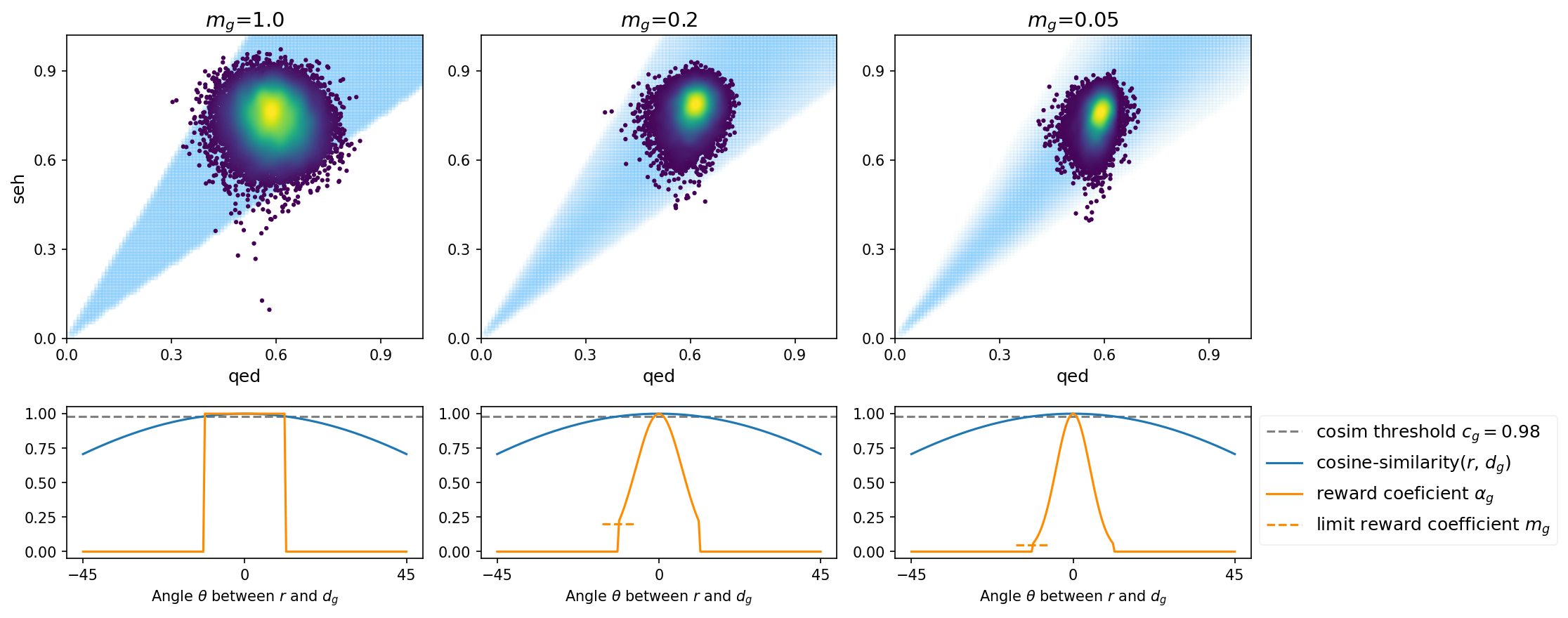}
    \caption{Effect of the hyperparameter $m_g$ on the profile of the reward coefficient $\alpha_g$ and the learned sampling distribution (top row).}
    \label{fig:focus_limit_coef}
\end{figure}

While the GFN model is given the goal direction $d_g$ as input, the width of the goal region, which depends on the cosine-similarity threshold $c_g$ is fixed, and the model adapts to producing samples within the region over time by trial-and-error. One can trade off the level of controllability of the goal-conditioned model with the difficulty of reaching those goals by increasing or reducing $c_g$. Another way to increase the controllability \textit{and} goal-reaching accuracy without drastically affecting the difficulty of reaching such goals is to make the model preferentially generate samples near the center of the focus region, thus reducing the risk of producing an out-of-focus sample due to epistemic uncertainty. To do so, we modify Equation~\ref{eq:goal-reward} and add a reward-coefficient $\alpha_g$, which further modulates the magnitude of the scalar reward based on how close to the center of the focus region the sample was generated. While many shaping functions could be devised, we choose the following form:
\begin{equation}
\label{eq:goal-reward-extended}
    R_g(x) = 
    \begin{cases}
    \alpha_g \sum_k r_k ,& \text{if } r \in g\\
    0, & \text{otherwise}
    \end{cases}
    \quad,\qquad \quad \alpha_g = \left( \frac{r \cdot d_g}{||r||\cdot||d_g||} \right) ^{\frac{\log m_g}{\log c_g}}
\end{equation}
In words, the reward coefficient $\alpha_g$ is equal to the cosine similarity between the reward vector $r$ and the goal direction $d_g$ exponentiated in such a way that $\alpha_g=m_g$ at the limit of the focus region. So for example, setting $m_g=0.2$ means that the reward is maximal at the center of the focus region, is at 20\% of that magnitude at the limit of the focus region, and follows a sharp sigmoid-like profile in between. Figure~\ref{fig:focus_limit_coef} showcases the reward coefficient as a function of the angle between $r$ and $d_g$ for different values of $m_g$ and the corresponding distributions learned by the model. We can see that a smaller value of $m_g$ encourages the model to produce samples in a more focused way towards the center of the goal region. Importantly, with a large enough value of $m_g$, this design preserves the notion of a well-defined goal region (positive reward inside the region and zero reward outside) and thus also preserves our ability to reason about goal-reaching accuracy, a beneficial concept for monitoring the model, filtering out-of-focus samples, etc.

\clearpage
\subsection{Tabular Goal-Sampler}
\label{app:learned_goal_model}

To cope with the problem of infeasible goal regions described in Section~\ref{sec:learned_goal_distribution}, we explore the idea of sampling the goal directions $d_g$ from a learned goal distribution rather than sampling all directions uniformly. The idea is that, as the model learns about which goal directions point towards infeasible regions of the objective space, we can attribute a much lower sampling likelihood to these regions in order to focus on more fruitful goals. 

We implement a first version of this idea as a tabular goal-sampler (Tab-GS). We first build a dataset of goal directions $\mathcal{D}_G$. This could be done in many different ways such as sampling a large number of positive vectors at the surface of the unit hypersphere in objective space. In our case, we discretise the extreme faces of the unit hypercube and normalize them. At training time, for each direction vector $d_g \in \mathcal{D}_G$, we keep a count of the number of samples which have landed closest to it (closer than any other direction $d_g'$) and follow this very simple scheme: from the beginning up to 25\% of the training iterations, we sample batches of goal directions $\{d_g\}_{i=1}^N$ uniformly over $\mathcal{D}_G$. Then starting at 25\% of the training iterations, while we keep updating each direction's count, we sample batches of goal directions according to the following (unnormalized) likelihoods:

\begin{equation}
\label{eq:goal-sampling-likelihood}
    f(d_g) = 
    \begin{cases}
    1 & \text{if } d_g \text{ has never been sampled}\\
    1 & \text{if there has been a sample } r \text{ closest to } d_g \text{ than any other goal direction in } \mathcal{D}_G\\
    0.1 & \text{otherwise}
    \end{cases}
\end{equation}

Finally, at 75\% of the training, we stop updating the goal direction counts to allow the model to fine-tune itself to a now stationary goal-distribution Tab-GS($f$). At test time we also sample from that same stationary distribution. 

\begin{figure}[h!]
    \centering
    \includegraphics[width=1.\textwidth]{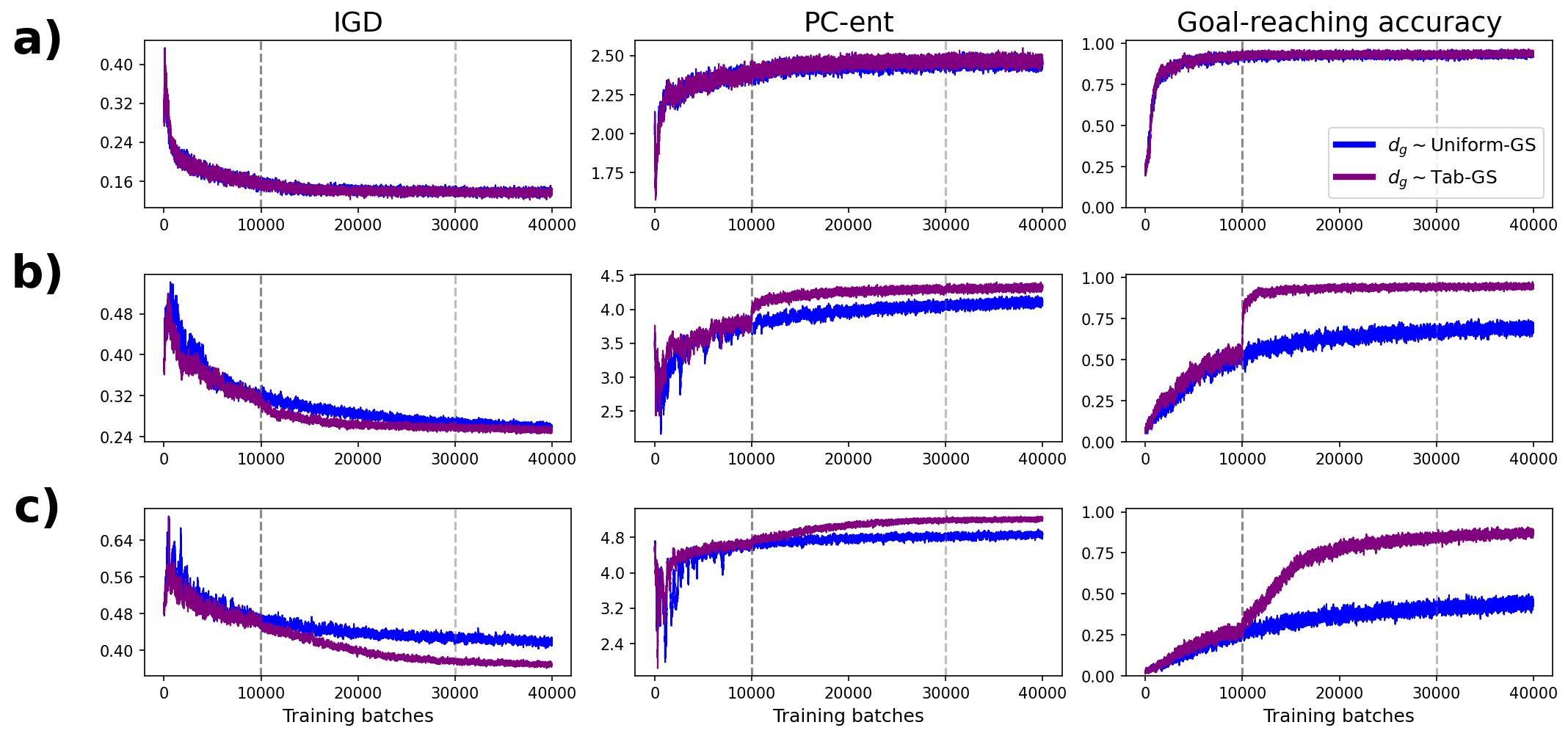}
    \caption{Learning curves for our goal-conditioned model trained by sampling goal directions $d_g$ either uniformly on the positive quadrant of a $K$-dimensional hypersphere (Uniform-GS) in blue or according to our learned tabular goal-sampler (Tab-GS) in purple on \textbf{a)} 2 objectives, \textbf{b)} 3 objectives and \textbf{c)} 4 objectives. Vertical dotted lines indicate 25\% and 75\% of training when we start sampling goal directions according to Equation~\ref{eq:goal-sampling-likelihood} and when we stop updating the learned goal-sampler, respectively.}
    \label{fig:effect_of_tabgs}
\end{figure}

Figure~\ref{fig:effect_of_tabgs} shows the effect of our learned tabular goal-sampler (Tab-GS) on the model's performance and learning dynamics. While the 2-objective problem does not contain a lot of infeasible goal directions, resulting in very similar behaviors for both methods, we can see that in the case of 3 and 4 objectives, the model experiences an important immediate improvement in goal-reaching accuracy at 25\% of training when we start sampling $d_g$'s according to our learned goal-sampler and that this improved focus helps the model further improves on these more fruitful goal directions, resulting in an increase IGD and PC-ent scores.

\clearpage
\section{Additional Results}
\label{app:additional_results}

\begin{figure}[h!]
    \centering
    \begin{minipage}{.5\textwidth}
        \centering
        \includegraphics[width=\textwidth]{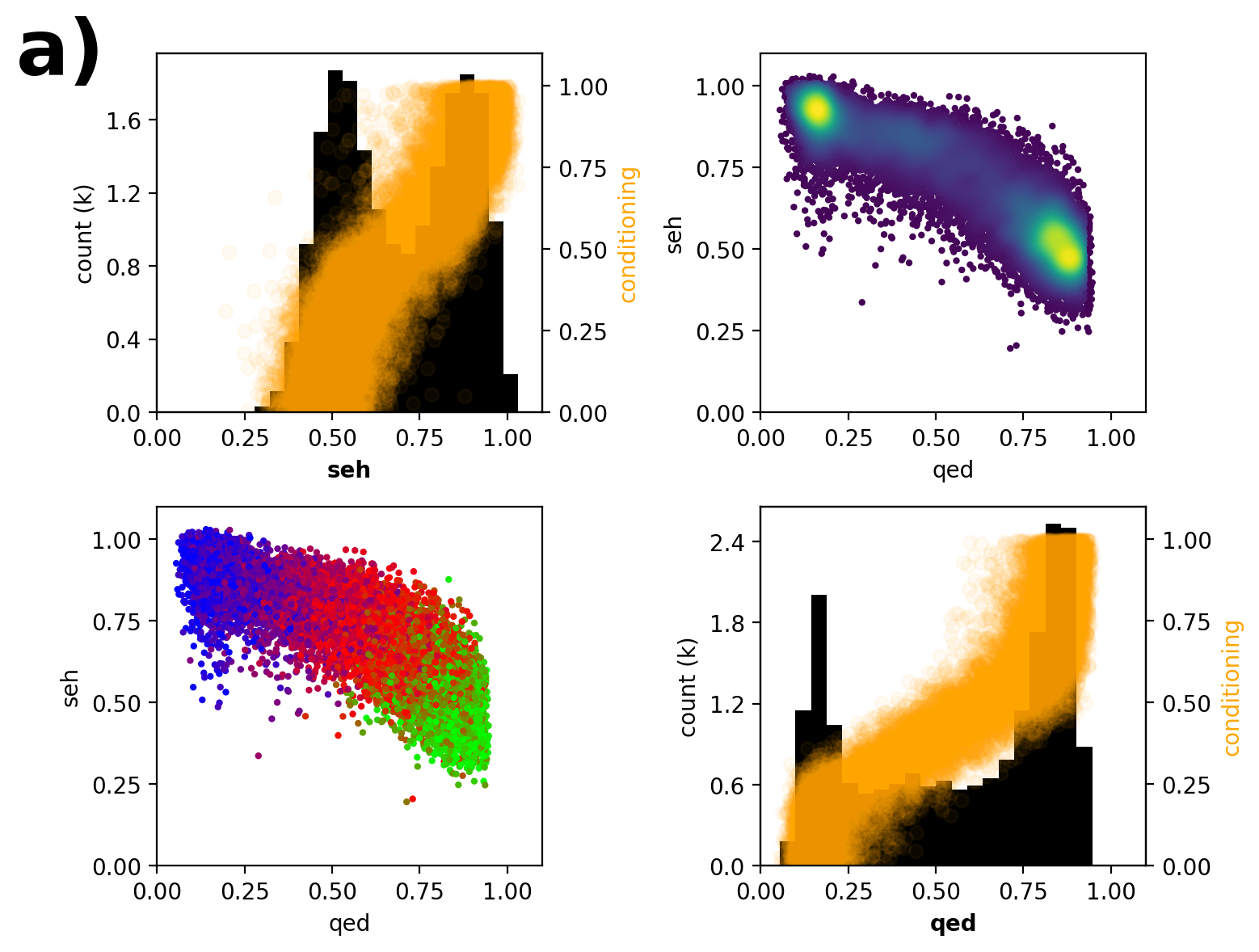}
    \end{minipage}%
    \begin{minipage}{0.5\textwidth}
        \centering
        \includegraphics[width=\textwidth]{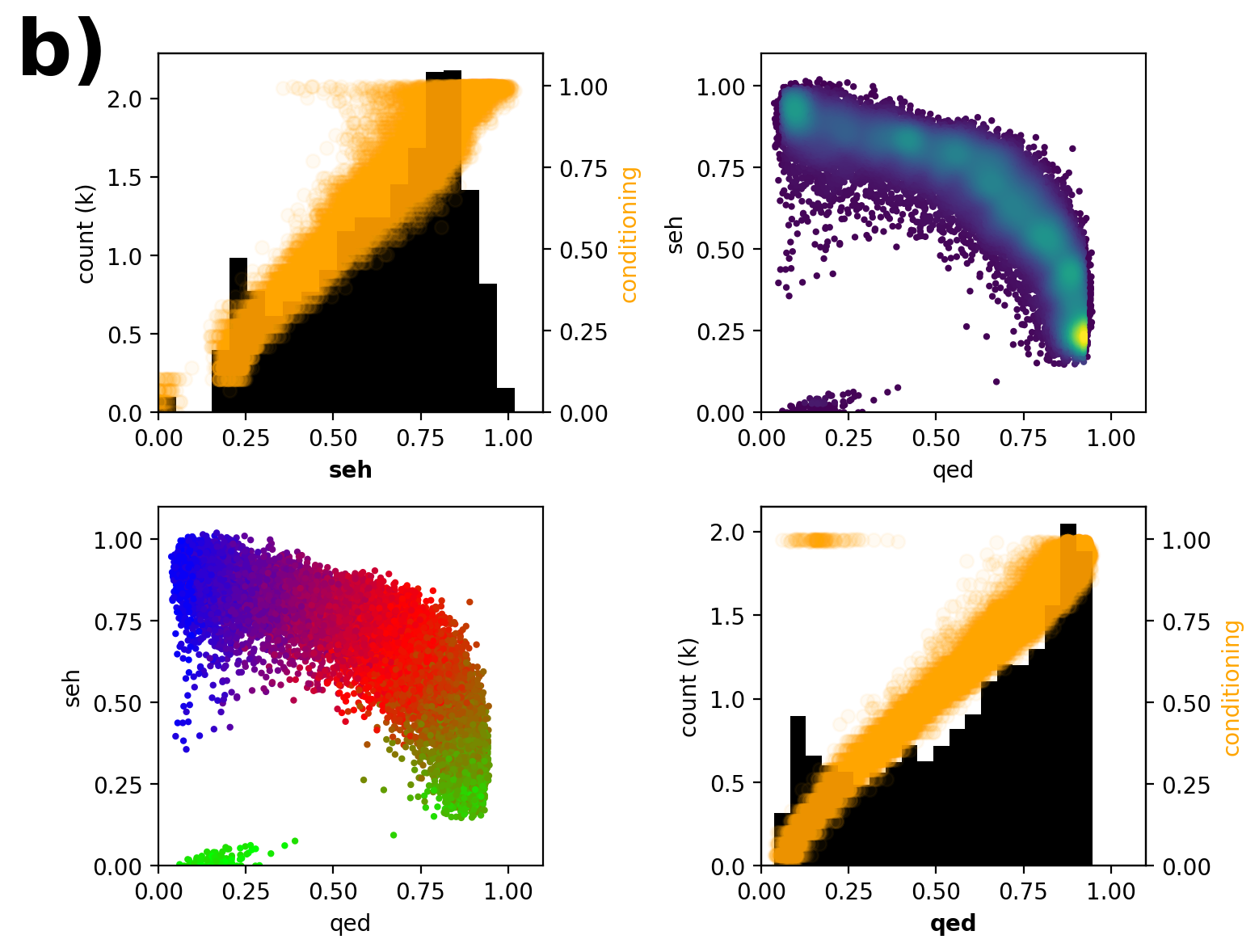}
    \end{minipage}
    \caption{Comparison between \textbf{a)} preference-conditioned and \textbf{b)} goal-conditioned models trained on the 2-objective problem (seh, qed). Each panel is an assemblage of $K \times K$ plots where $K$ is the number of objectives. \textbf{On the diagonal}, each plot focuses on a single objective. They each show a histogram (dark) of the samples' scores $r_{\cdot,k}$ for that objective, overlayed with a scatter plot (orange) in which each point is a distinct sample $i$ with coordinates $(x,y)=(r_{i,k},c_{i,k})$, where $r_{i,k}$ is the reward attributed to sample $i$ for objective $k$ and $c_{i,k}$ is the corresponding value of the conditioning vector that was used to generate that sample. The histogram thus showcases the distribution and span of our set of samples for a given dimension in objective space while the scatter plot allows us to visualise the correlation between the conditioning vectors and the resulting rewards for that dimension. \textbf{Above the diagonal}, each plot shows the density of the learned distribution on the plane corresponding to a pair of objectives. Brighter colors indicate that a region is more densely populated. \textbf{Below the diagonal}, each plot shows the controllability of the learned distribution where BRG colors represent the angle between the vector $[1, 0]$ and \textbf{a)} the preference-vector $w$ or \textbf{b)} the goal direction $d_g$ (this is the same as in Figure~\ref{fig:difficult_pareto_fronts_alignment}). Overall, we can see that on the density plot and on the histograms that the goal-conditioned approach produces a more uniformly distributed set of samples while the orange scatter plot and the BRG-colored plots show that they also provide a finer control over the generated samples.}
    \label{fig:all_plots_2_objectives}
\end{figure}

\begin{figure}[h!]
    \centering
    \begin{minipage}{.5\textwidth}
        \centering
        \includegraphics[width=\textwidth]{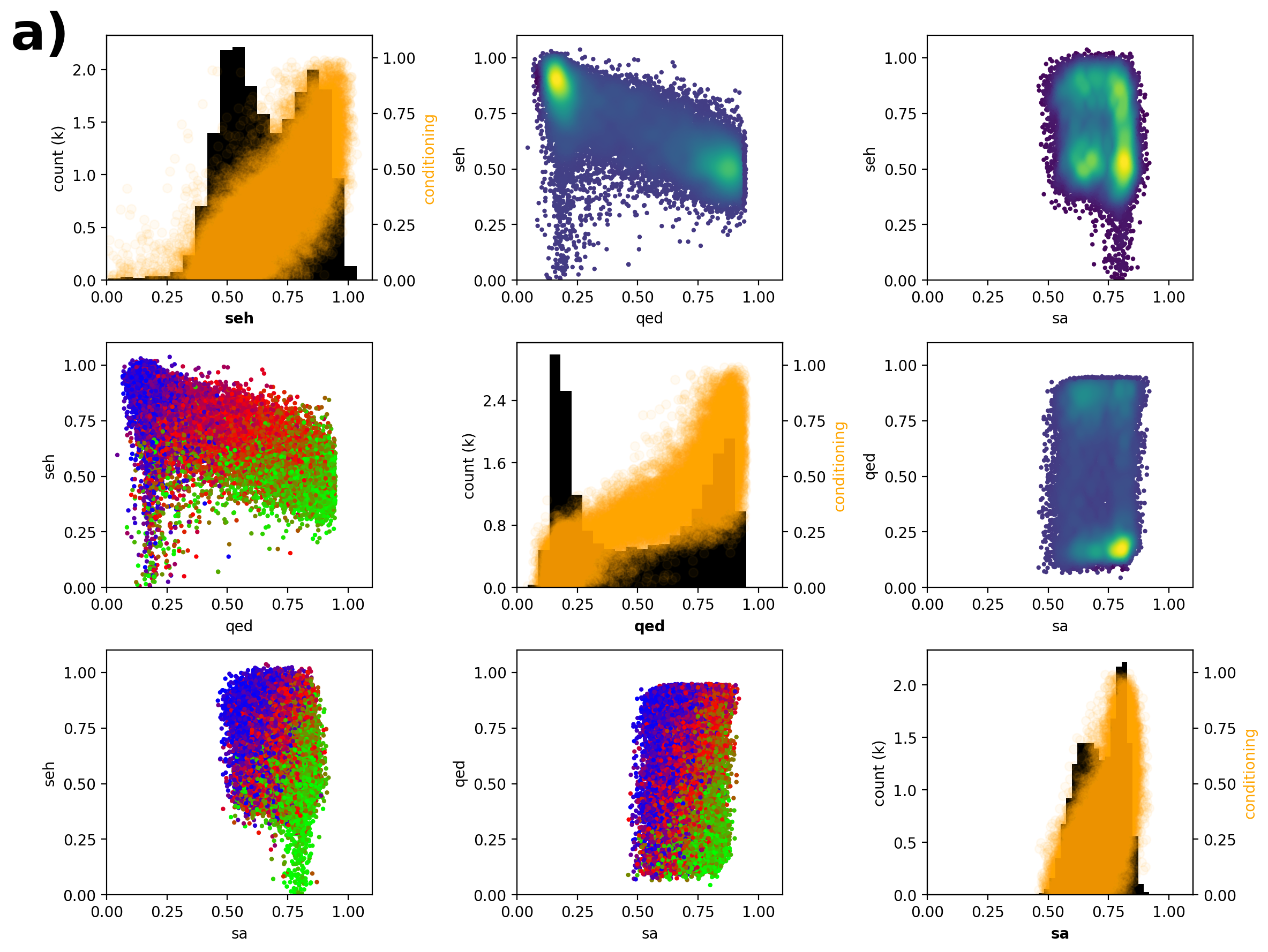}
    \end{minipage}%
    \begin{minipage}{0.5\textwidth}
        \centering
        \includegraphics[width=\textwidth]{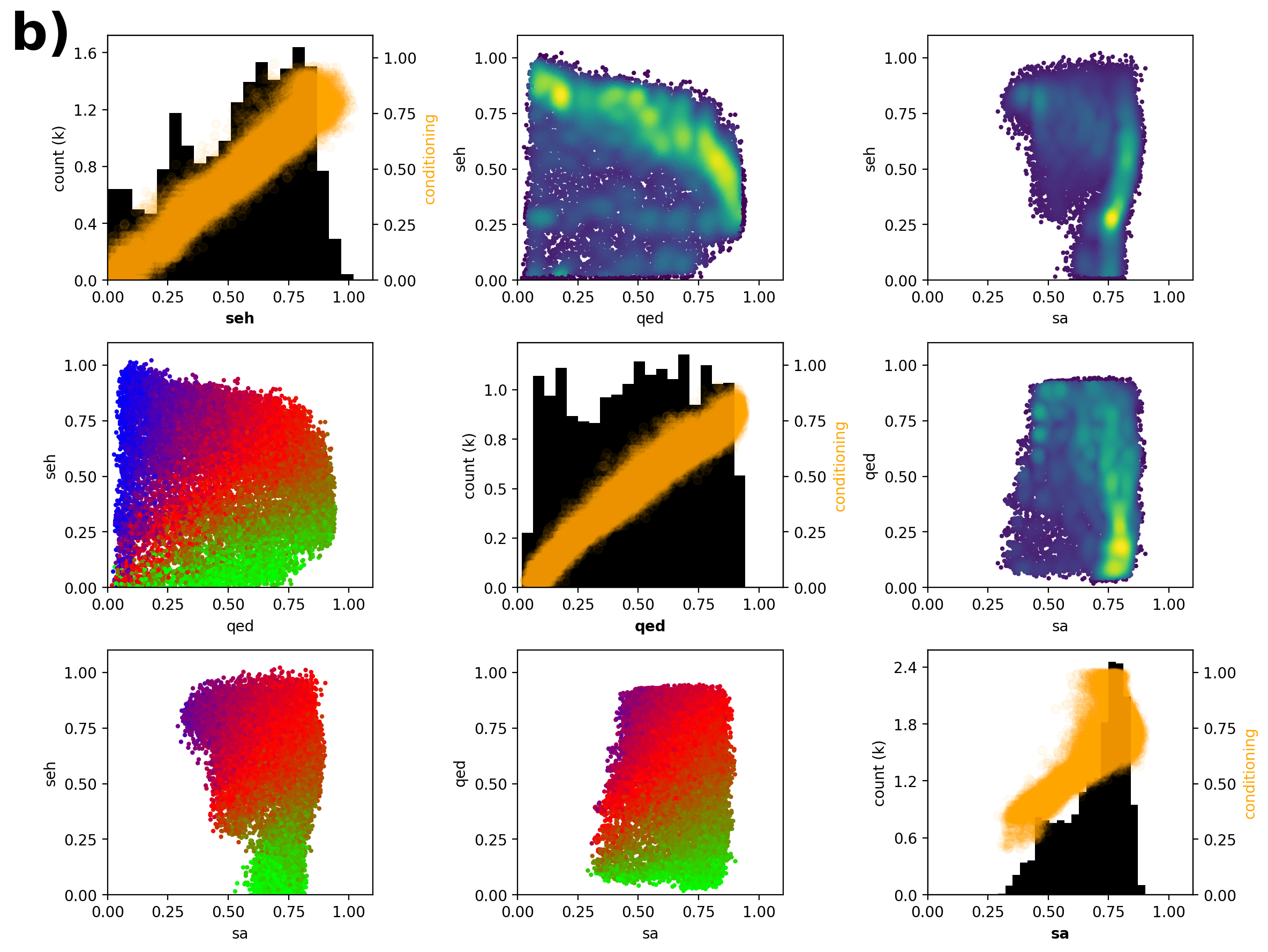}
    \end{minipage}
    \caption{Idem to Figure~\ref{fig:all_plots_3_objectives} but with 3 objectives: seh, qed, sa.}
    \label{fig:all_plots_3_objectives}
\end{figure}

\begin{figure}[h!]
    \centering
    \includegraphics[width=0.85\textwidth]{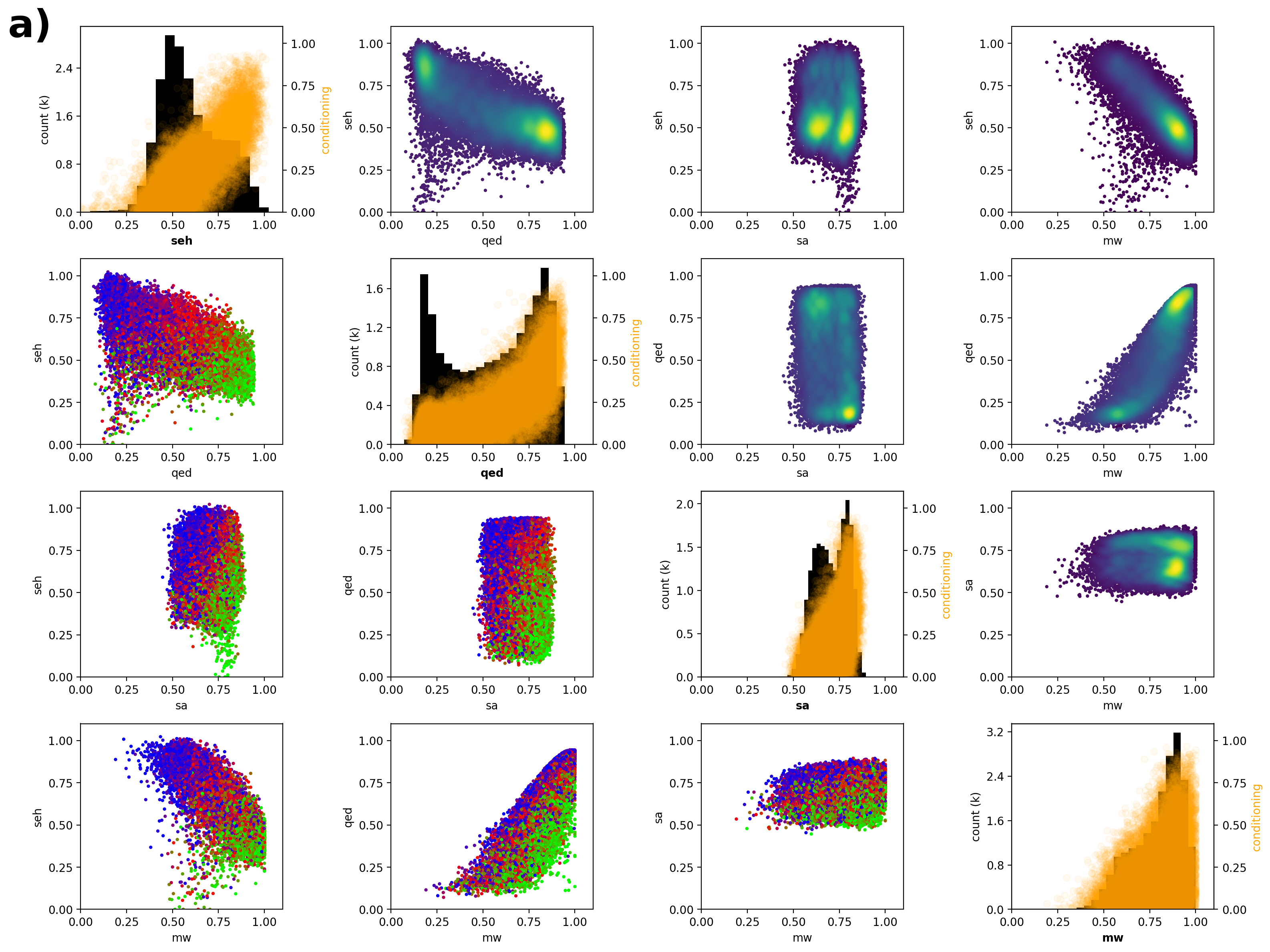}
    \includegraphics[width=0.85\textwidth]{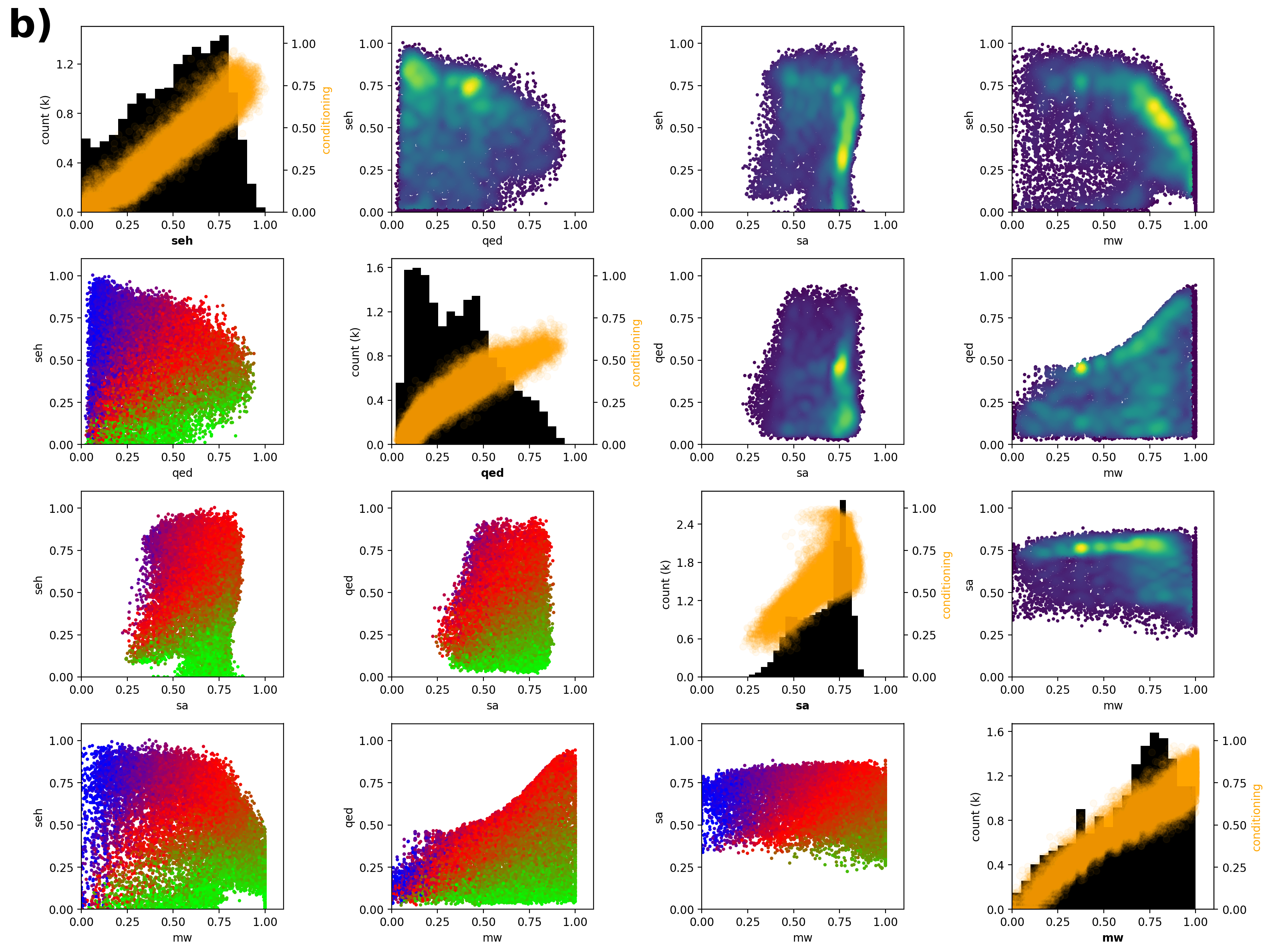}
    \caption{Idem to Figure~\ref{fig:all_plots_2_objectives} but with 4 objectives: seh, qed, sa, mw.}
    \label{fig:all_plots_4_objectives}
\end{figure}

In this section, we present additional plots for experiments on 2, 3 and 4 objectives (Figures~\ref{fig:all_plots_2_objectives}, \ref{fig:all_plots_3_objectives} \& \ref{fig:all_plots_4_objectives}).


%%%%%%%%%%%%%%%%%%%%%%%%%%%%%%%%%%%%%%%%%%%%%%%%%%%%%%%%%%%%%%%%%%%%%%%%%%%%%%%
%%%%%%%%%%%%%%%%%%%%%%%%%%%%%%%%%%%%%%%%%%%%%%%%%%%%%%%%%%%%%%%%%%%%%%%%%%%%%%%

\end{document}